# AI-driven Automation of End-to-end Assessment of Suturing Expertise

Authors: Atharva Deo[1], Nicholas Matsumoto[1], Sun Kim[1], Peter Wager[1], Randy G. Tsai[2], Aaron Denmark[1], Cherine Yang[1], Xi Li[1], Jay Moran[1], Miguel Hernandez[1], Andrew J. Hung[1]

Affiliations:

1. Cedars Sinai Medical Center, Los Angeles, California

2. Universityof California Los Angeles, California

Key information:

1. Research question: Can we automate the end-to-end assessment of suturing expertise, and what benefits would it offer?
2. Findings: All the models performed well with an AUC >= 0.75, indicating robust and high prediction performance.
3. Meaning: The AI-driven approach can be used to automate scoring, vastly reducing the skills assessment time and enabling real-time feedback delivery to surgeons and trainees.

MANUSCRIPT

Introduction

We present an AI based approach to automate the End-to-end Assessment of Suturing Expertise (EASE), a suturing skills assessment tool that comprehensively defines criteria around relevant sub-skills.[1] While EASE provides granular skills assessment related to suturing to provide trainees with an objective evaluation of their aptitude along with actionable insights, the scoring process is currently performed by human evaluators, which is time and resource consuming. The AI based approach solves this by enabling real-time score prediction with minimal resources during model inference. This enables the possibility of real-time feedback to the surgeons/trainees, potentially accelerating the learning process for the suturing task and mitigating critical errors during the surgery, improving patient outcomes. In this study, we focus on the following 7 EASE domains that come under 3 suturing phases: 1) Needle Handling: Number of Repositions, Needle Hold Depth, Needle Hold Ratio, and Needle Hold Angle; 2) Needle Driving: Driving Smoothness, and Wrist Rotation; 3) Needle Withdrawal: Wrist Rotation.

Material and methods

The data comprised of 80 suturing videos, each with 24 stitches, totalling 13,440 video datapoints. Multiscale Vision Transformer (MViTv2) model[2] was used to predict binary EASE scores for four domains: Needle Handling (Number of Repositions), Needle Driving (Driving Smoothness, Wrist Rotation), and Needle Withdrawal (Wrist Rotation). The MViTv2 model processed 16 frames of 224×224 resolution. A 3D-CNN model with three Conv3D-ReLU-MaxPool3D layers was used for Needle Handling (Hold Depth, Hold Ratio, Hold Angle), processing 16 frames of 384×384 resolution. The 3D-CNN was chosen due to its superior performance on these domains with shorter temporal dependencies. A fully connected layer (FC layer) was used to process the spatiotemporal features extracted by the models (Figure 1a). A model routing approach selected the appropriate model based on the suturing skill domain under focus as a router (Figure 1b). The best performing model was tested on the test data using 80:20 training:test split on the data. Data augmentation strategies such as Resizing, Rotation, Horizontal, Vertical Flip was used to improve the model performance and generalization capabilities.

Results

All the models perform well, above an AUC of 0.75 (Table 1), indicating a strong and robust prediction performance. The best-performing model was for Needle Handling: Number of Repositions suturing domain, achieving an AUC of 0.82 (95% CI: 0.78 - 0.86), which could be attributed to its clearly distinguishable temporal action. The model for the Needle Handling: Hold Ratio suturing domain performed the worst in terms of confidence interval lower bound, achieving an AUC of 0.79 (95% CI: 0.66 - 0.92), which could be due to image occlusions, including varying orientations of the robotic jaw relative to the needle. Such variations introduce significant diversity in the training images, potentially reducing prediction accuracy.

Discussion and Conclusion

We developed AI models with strong and robust prediction capabilities that can automate the EASE scoring process, leading to fast, real-time predictions that also consume less resources during model inference. While the models perform well, currently they require the suturing domain input to use the appropriate model to predict the EASE score. To fully automate the EASE scoring process, we propose a

2-stage pipeline as our future work, where the first stage predicts the suturing domain using the video input, which is fed to the second stage, which we have already developed, along with the video input to predict the EASE score.

Disclosures

Research reported in this publication was supported by National Cancer Institute of the National Institutes of Health under award number R01CA251579. The content is solely the responsibility of the authors and does not necessarily represent the official views of the National Institutes of Health.

*APPENDIX*

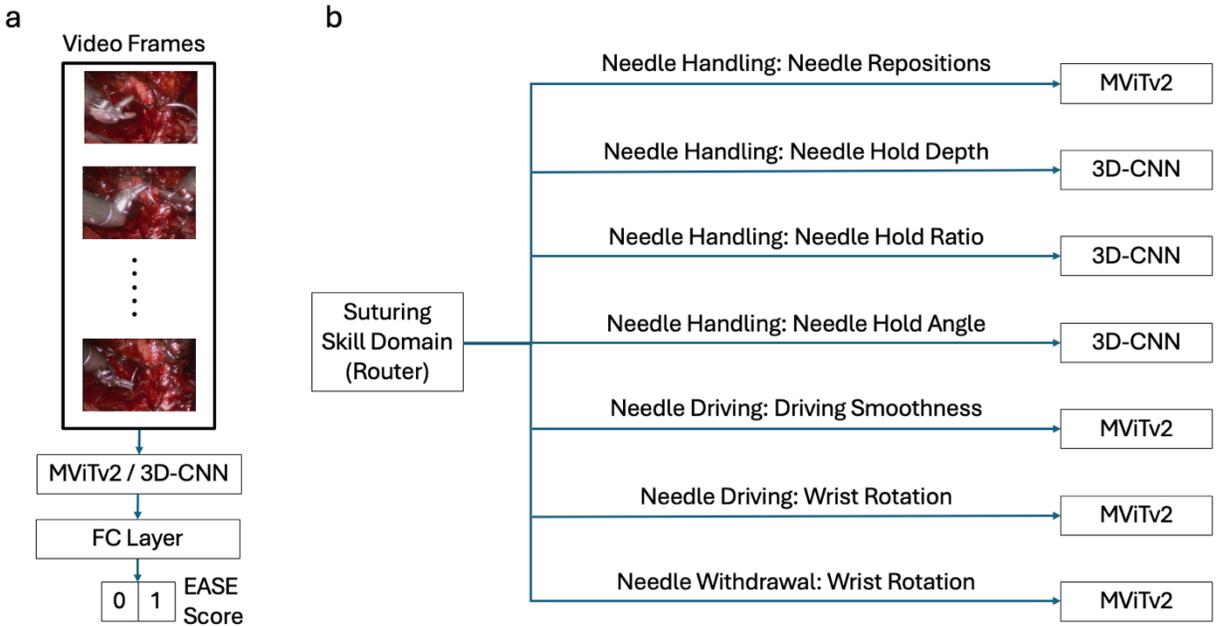

Figure 1: a) EASE score prediction pipeline. Video frames are processed by the MViTv2/ 3D-CNN model to extract spatiotemporal features which is processed by the FC layer to predict binary score. b) Model routing-based architecture for EASE score prediction across different suturing phases and skill domains.

Table 1: Model performance across different suturing phases and skill domains

| Suturing Phase: Skill Domain | AUC Scores [95% CI] |
|---|---|
| Needle Handling: Number of Repositions | 0.82 [0.78, 0.86] |
| Needle Handling: Needle Hold Depth | 0.82 [0.73, 0.90] |
| Needle Handling: Needle Hold Ratio | 0.79 [0.66, 0.92] |
| Needle Handling: Needle Hold Angle | 0.79 [0.69, 0.88] |
| Needle Driving: Driving Smoothness | 0.78 [0.73, 0.83] |
| Needle Driving: Wrist Rotation | 0.75 [0.70, 0.80] |
| Needle Withdrawal: Wrist Rotation | 0.76 [0.67, 0.85] |